%% file: main.tex
\documentclass{article}

\PassOptionsToPackage{numbers, compress}{natbib}
\usepackage[preprint]{neurips_2026}

\usepackage[utf8]{inputenc}
\usepackage[T1]{fontenc}
\usepackage{hyperref}
\usepackage{url}
\usepackage{booktabs}
\usepackage{array}
\usepackage{amsfonts}
\usepackage{nicefrac}
\usepackage{microtype}
\usepackage{xcolor}
\usepackage{enumitem}
\usepackage{algorithm}
\usepackage[noend]{algorithmic}
\usepackage{amsmath}
\usepackage{graphicx}
\usepackage{amssymb}
\usepackage{multirow}
\usepackage{subcaption}
\usepackage{wrapfig}
\usepackage[table]{xcolor}
\usepackage{tcolorbox}
\tcbuselibrary{breakable, skins}

\newcommand{\ours}{\textsc{EvolveMem}}

\title{\ours{}: Self-Evolving Memory Architecture via AutoResearch for LLM Agents}

\author{
  Jiaqi Liu$^{1}$ \quad
  Xinyu Ye$^{1}$\thanks{Corresponding Authors} \quad
  Peng Xia$^{1}$ \quad
  Zeyu Zheng$^{2}$ \\[2pt]
  \textbf{Cihang Xie}$^{3}$ \quad
  \textbf{Mingyu Ding}$^{1}$ \quad
  \textbf{Huaxiu Yao}$^{1}$\footnotemark[1] \\[4pt]
  $^{1}$UNC-Chapel Hill \quad
  $^{2}$UC Berkeley \quad
  $^{3}$UCSC
}

\begin{document}

\maketitle

\input{section/abstract}

\input{section/intro}
\input{section/related_work}

\input{section/method}
\input{section/experiments}

\input{section/conclusion}

\bibliographystyle{plainnat}
\bibliography{main}

\appendix
\input{section/appendix}

\end{document}

%% file: section/abstract.tex
\begin{abstract}
Long-term memory is essential for LLM agents that operate across multiple sessions, yet existing memory systems treat retrieval infrastructure as fixed: stored content evolves while scoring functions, fusion strategies, and answer-generation policies remain frozen at deployment. We argue that truly adaptive memory requires co-evolution at two levels: the stored knowledge and the retrieval mechanism that queries it. We present \ours{}, a self-evolving memory architecture that exposes its full retrieval configuration as a structured action space optimized by an LLM-powered diagnosis module. In each evolution round, the module reads per-question failure logs, identifies root causes, and proposes targeted configuration adjustments; a guarded meta-analyzer applies them with automatic revert-on-regression and explore-on-stagnation safeguards. This closed-loop self-evolution realizes an AutoResearch process: the system autonomously conducts iterative research cycles on its own architecture, replacing manual configuration tuning. Starting from a minimal baseline, the process converges autonomously, discovering effective retrieval strategies including entirely new configuration dimensions not present in the original action space. On LoCoMo, \ours{} outperforms the strongest baseline by 25.7\% relative and achieves a 78.0\% relative improvement over the minimal baseline. On MemBench, \ours{} exceeds the strongest baseline by 18.9\% relative. Evolved configurations transfer across benchmarks with positive rather than catastrophic transfer, indicating that the self-evolution process captures universal retrieval principles rather than benchmark-specific heuristics. Code is available at \url{https://github.com/aiming-lab/SimpleMem}.
\end{abstract}

%% file: section/intro.tex
\section{Introduction}
\label{sec:intro}

Persistent memory is a foundational capability for long-running LLM agents. Personal assistants must remember user preferences across months; coding agents must track evolving project decisions; customer-facing systems must maintain coherent identities across sessions~\citep{zhang2024survey, packer2023memgpt, sumers2024coala}. These scenarios require memory systems that grow with the agent, but growth introduces a problem that has been largely overlooked: as the scale and complexity of stored memories change, the retrieval strategy stays the same. Different types of questions fundamentally require different retrieval strategies: factual lookups need precise keyword matching, temporal reasoning needs time-aware filtering, multi-hop inference needs query decomposition. A frozen retrieval configuration cannot optimally serve all of these needs simultaneously.

Recent memory architectures have advanced along two fronts. One line focuses on memory organization: MemGPT~\citep{packer2023memgpt} manages working and long-term memory through tiered storage, Mem0~\citep{chhikara2025mem0} and A-MEM~\citep{xu2025amem} structure memory content with knowledge graphs and associative networks, and SimpleMem~\citep{liu2026simplemem} compresses conversations into retrieval-friendly units. Another line focuses on memory maintenance: MemoryBank~\citep{zhong2024memorybank} applies forgetting curves to prune stale entries, and various consolidation mechanisms deduplicate and merge redundant information. Despite their diversity, all these systems share a fundamental assumption: the memory content evolves over time, but the retrieval infrastructure remains frozen. Scoring functions, fusion weights, context budgets, and answer-generation strategies stay unchanged throughout the agent's lifetime.

\begin{figure}[t]
    \centering
    \includegraphics[width=\textwidth]{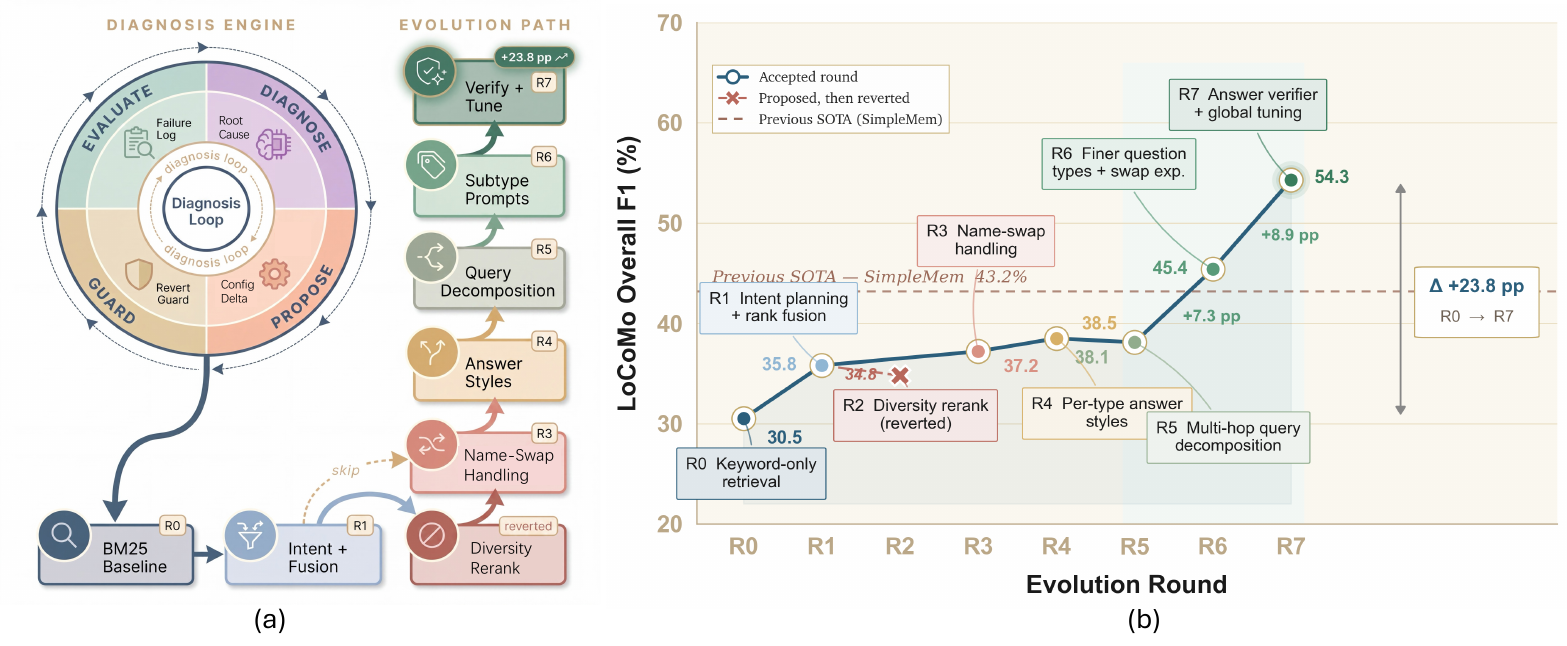}
    \caption{\textbf{\ours{} self-evolves its retrieval configuration on LoCoMo via AutoResearch.}
    \textbf{(a)} A four-step evolution loop (\textsc{Evaluate}--\textsc{Diagnose}--\textsc{Propose}--\textsc{Guard}) ratchets accepted proposals into the action space; harmful ones (e.g., R2) are auto-reverted.
    \textbf{(b)} Overall F1 trajectory (single-backbone GPT-4o): 30.5\% baseline to 54.3\% at R7.}
    \label{fig:teaser}
    \vspace{-1.5em}
\end{figure}

This assumption creates a mismatch that worsens over time. As stored memories grow from dozens to hundreds of heterogeneous records, a retrieval policy calibrated for the small store becomes suboptimal, and different question categories require fundamentally different retrieval strategies. Our key observation is that a truly adaptive memory system must evolve at two levels: the stored knowledge must be maintained and consolidated, and the retrieval infrastructure itself must self-adapt to the changing memory landscape and query distribution. Achieving such self-adaptation requires the system to autonomously observe its own failures, hypothesize root causes, test configuration changes, and retain only those that improve performance.

We present \ours{}, a memory architecture that autonomously evolves its retrieval infrastructure through LLM-driven closed-loop diagnosis. \ours{} combines a typed knowledge store with a multi-view retriever covering lexical, semantic, and structured-metadata signals, and exposes the complete retrieval configuration as a structured action space. An LLM-powered diagnosis module reads per-question failure logs, categorizes root causes, and proposes targeted configuration adjustments that a guarded meta-analyzer applies with automatic revert-on-regression safeguards. This closed-loop self-evolution constitutes an AutoResearch process: the system autonomously conducts the observe-hypothesize-experiment-validate cycle that would otherwise require manual researcher effort, discovering effective retrieval policies including entirely new configuration dimensions not present in the original framework.

In summary, our primary contribution is \ours{}, the first memory framework that autonomously evolves its retrieval infrastructure through LLM-driven closed-loop diagnosis, realizing an AutoResearch process that replaces manual configuration tuning. On LoCoMo, \ours{} outperforms the strongest published baseline by 25.7\% relative (78.0\% over the minimal baseline); on MemBench, it exceeds the strongest baseline by 18.9\% relative. The evolved configurations transfer across benchmarks with positive rather than catastrophic transfer.

%% file: section/related_work.tex
\section{Related Work}
\label{sec:related}

\noindent \textbf{Memory systems for LLM agents.}
Persistent memory has become a core component of LLM agent architectures \citep{hu2025memory, zhang2024survey, sumers2024coala, wei2025evomemory}. Reflexion \citep{shinn2023reflexion} and Generative Agents \citep{park2023generative} maintain episodic buffers indexed by recency and importance. MemGPT \citep{packer2023memgpt} introduces OS-inspired tiered memory; MemoryBank \citep{zhong2024memorybank} applies Ebbinghaus-inspired forgetting. SCM \citep{wang2024scm} extracts entity-aware summaries; Mem0 \citep{chhikara2025mem0} builds knowledge graphs; A-MEM \citep{xu2025amem} creates Zettelkasten-style networks; MemSkill \citep{zhang2026memskill} evolves reusable memory skills. SimpleMem \citep{liu2026simplemem, omni_simplemem2026}, SeCom \citep{pan2025secom}, and RMM \citep{tan2025rmm} address retrieval quality through semantic compression, topic-level segmentation, and reflective refinement respectively. LongMem \citep{wang2023longmem} and MemoryLLM \citep{wang2024memoryllm} embed long-term knowledge directly into model parameters. All these systems evolve stored \emph{content} but keep the retrieval infrastructure frozen. \ours{} addresses this gap by making the full retrieval configuration self-evolving via LLM-powered closed-loop diagnosis, an approach we characterize as AutoResearch applied to the system's own architecture. Table~\ref{tab:comparison} summarizes key architectural differences.

\begin{wraptable}{r}{0.55\textwidth}
  \centering
  \small       
  \vspace{-1em}
  \caption{Comparison of memory systems. \ours{} is the first to combine content evolution with self-evolving
  retrieval infrastructure via AutoResearch. \textbf{C.E.}: content evolution. \textbf{P.E.}: policy/parameter evolution.
  \textbf{T.M.}: typed memory. \textbf{Cons.}: consolidation. \textbf{O.E.}: offline evaluation.}
  \vspace{-0.5em} 
  \label{tab:comparison}                                                
  \setlength{\tabcolsep}{4pt}
  \begin{tabular}{@{}lccccc@{}}
  \toprule                                           
  \textbf{System} & \textbf{C.E.} & \textbf{P.E.} & \textbf{T.M.} & \textbf{Cons.} & \textbf{O.E.} \\
  \midrule                                                
  MemGPT \citep{packer2023memgpt} & \checkmark & & & \checkmark & \\
  SCM \citep{wang2024scm} & \checkmark & & & \checkmark & \\       
  SimpleMem \citep{liu2026simplemem} & \checkmark & & & \checkmark & \\
  Mem0 \citep{chhikara2025mem0} & \checkmark & & & \checkmark & \\
  A-MEM \citep{xu2025amem} & \checkmark & & & \checkmark & \\     
  Memory-R1 \citep{liang2025memoryr1} & \checkmark & & & & \\ 
  Agentic Memory \citep{yu2026agenticmemory} & \checkmark & & \checkmark & & \\                                       
  MemoryBank \citep{zhong2024memorybank} & \checkmark & & & \checkmark & \\                                           
  MemEvolve \citep{zhang2025memevolve} & \checkmark & \checkmark & & & \\
  \midrule                                   
  \ours{} & \checkmark & \checkmark & \checkmark & \checkmark & \checkmark \\
  \bottomrule  
  \end{tabular}                                                   
  \vspace{-1.0em} 
  \end{wraptable}          
\noindent \textbf{Adaptive retrieval.}
RAG \citep{lewis2020retrieval, guu2020realm} enriches LLM inputs with external knowledge. Recent variants adapt \emph{when} and \emph{what} to retrieve: Self-RAG \citep{asai2024selfrag} uses reflection tokens, CRAG \citep{yan2024corrective} adds corrective quality checks, FLARE \citep{jiang2023active} triggers retrieval when generation confidence drops, and Adaptive-RAG \citep{jeong2024adaptiverag} routes queries by estimated complexity. LLM-powered database tuning \citep{giannakouris2025lambdatune} and reinforcement-learning-based index optimization \citep{wang2025litune} demonstrate that retrieval parameters can be auto-optimized from workload statistics. These approaches adapt retrieval triggers or post-retrieval filtering, but none adapts the retrieval \emph{parameters} (scoring weights, fusion mode, context budgets) over a deployed system's lifetime. \ours{} fills this gap through offline evolution over a structured action space.

\noindent \textbf{Self-improving agents and AutoResearch.}
Self-improvement has been explored via self-play \citep{chen2024self}, iterative refinement \citep{madaan2023self}, and evolutionary optimization \citep{gao2025survey}. Voyager \citep{wangvoyager} builds an expanding skill library; ExpeL \citep{zhao2024expel} extracts reusable insights from task trajectories; EvolveR \citep{wu2025evolver} closes an experience-driven evolution loop; SkillRL \citep{xia2026skillrl} evolves agents via recursive skill augmentation; MemRL \citep{zhang2026memrl} applies runtime RL to episodic memory; Memory-R1 \citep{liang2025memoryr1} applies RL to memory operations; Agentic Memory \citep{yu2026agenticmemory} optimizes memory management with GRPO. MemEvolve \citep{zhang2025memevolve} jointly evolves agent knowledge and memory architecture. AutoResearchClaw \citep{liu2026autoresearchclaw} demonstrates that LLMs can conduct fully autonomous research pipelines, executing the complete cycle of hypothesis generation, experimental design, and result interpretation without human intervention. \ours{} applies this AutoResearch paradigm to a specific and previously unexplored target: the system autonomously researches its own retrieval infrastructure through iterative diagnosis-driven evolution, discovering architectural improvements that would otherwise require manual researcher effort. Unlike prior self-improving agents that optimize behavioral policies or stored content, \ours{} targets the retrieval mechanism itself as the research subject. Our consolidation mechanisms draw on complementary learning systems theory \citep{mcclelland1995there} and Ebbinghaus forgetting \citep{ebbinghaus1885memory}.

%% file: section/method.tex
\section{\ours{}}
\label{sec:method}

\begin{figure*}[t]
    \centering
    \includegraphics[width=\textwidth]{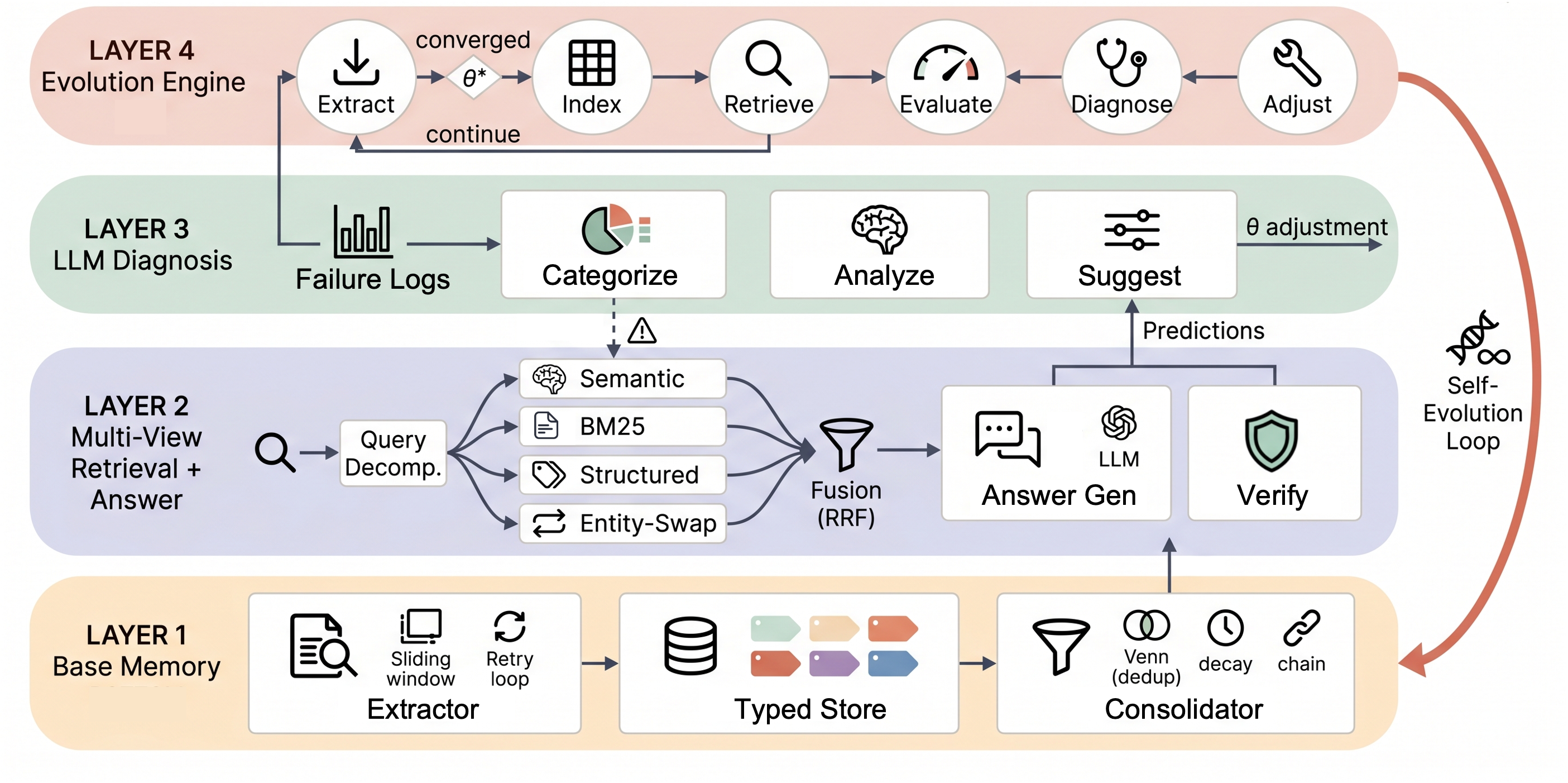}
    \caption{\textbf{\ours{} architecture.} Three layers connected by a self-evolution feedback loop. A typed memory store is populated by an LLM-based extractor with retry and chunk-splitting; a multi-view retriever fuses BM25, semantic, and structured-metadata search with optional entity-swap, query decomposition, and answer verification; an LLM-powered diagnosis module reads per-question raw-result logs and proposes structured adjustments to the retrieval configuration; the evolution engine validates adjustments and auto-converges when the primary metric plateaus.}
    \label{fig:architecture}
    \vspace{-1.5em}
\end{figure*}

The key design principle of \ours{} is that the retrieval infrastructure itself is a first-class optimization target, not a set of hand-tuned hyperparameters frozen at deployment time. Rather than relying on manual research to find good configurations, \ours{} automates the entire research process: it observes system behavior, diagnoses failure patterns, proposes architectural changes, and validates them empirically. As illustrated in Figure~\ref{fig:architecture}, three components realize this AutoResearch principle through a closed evolution loop. A \emph{Structured Memory Store} (\S\ref{sec:memory_layer}) builds and maintains a typed knowledge base through LLM-based extraction and consolidation. A \emph{Retrieval Layer} (\S\ref{sec:retrieval}) exposes its full configuration as an evolvable action space, enabling every parameter from fusion weights to answer generation style to be optimized jointly. A \emph{Self-Evolution Engine} (\S\ref{sec:evolution_engine}) closes the loop: it reads per-question failure logs, categorizes root causes, proposes targeted configuration adjustments, and applies them with safeguards against regression. This closed-loop self-evolution realizes an AutoResearch process, mirroring the observe-hypothesize-experiment-validate cycle of human research. Detailed formulations and threshold values are provided in Appendix~\ref{app:detailed_formulations}; the complete pipeline is given as Algorithm~\ref{alg:full_pipeline} in Appendix~\ref{app:full_algorithm}.

\subsection{Structured Memory Store}
\label{sec:memory_layer}

A self-evolving retrieval system is only as good as the memory it retrieves from. The memory layer provides a structured, high-quality knowledge base that supports multi-view retrieval across heterogeneous question types. This requires addressing three sub-problems: how to represent individual memories so that multiple retrieval views can operate over them, how to extract memories from raw conversations, and how to maintain store quality as memories accumulate over time.

\noindent \textbf{Memory representation.} Each memory unit is a tuple $m = (c,\; \mu,\; \mathbf{e},\; \boldsymbol{\eta}), \quad
\mu \in \mathcal{T},\; \mathbf{e} \in \mathbb{R}^d$, where $c$ is natural-language content, $\mathbf{e}$ is a dense embedding, $\mu$ is a memory type drawn from a six-category taxonomy $\mathcal{T}$ (covering episodic, semantic, preference, project state, working summary, and procedural knowledge), and $\boldsymbol{\eta}$ collects auxiliary metadata including importance, confidence, entity-reinforcement score, extracted entities (including persons and locations), topics, and creation timestamp.

\noindent \textbf{Memory extraction.} Given a source conversation $S = (u_1, \ldots, u_T)$, a sliding window of length $W$ partitions $S$ into overlapping segments. For each window, the extractor invokes the backbone LLM to produce a set of typed memory units, with context from the previous window to avoid duplication. Three mechanisms handle common failure modes during extraction. First, when an LLM call fails, the system retries with increasing wait intervals, preserving any partially extracted results. Second, when a window exceeds the LLM's context limit, the system splits it into smaller sub-windows and merges their outputs. Third, a coverage verifier compares extracted memories against reference keywords from the source text and triggers re-extraction for any missing content. Together, these mechanisms substantially improve extraction coverage.

\noindent \textbf{Consolidation.} Three lightweight passes maintain store quality. First, deduplication merges any pair $(m_i, m_j)$ whose Jaccard similarity over tokenized content exceeds a threshold $\tau_J$, retaining the higher-importance unit. Second, importance decay applies a linear schedule that reduces $\iota_i$ by a fixed rate $\alpha_d$ per time unit, with a floor $\iota_{\min}$ to prevent useful memories from vanishing entirely. Third, entity reinforcement increments $\rho_i$ by $\delta_\rho$ each time a memory's extracted entities co-occur with a new query, capped at $\rho_{\max}$. Both $\iota_i$ and $\rho_i$ are carried forward as part of the memory metadata $\boldsymbol{\eta}$ and enter the retrieval ranking in \S\ref{sec:retrieval}.

\subsection{Retrieval as an Evolvable Action Space}
\label{sec:retrieval}

The central insight of \ours{} is that retrieval configuration should not be a static set of hand-tuned parameters but a structured action space that evolves alongside the memory store. Different question types fundamentally require different retrieval strategies: factual lookups need exact keyword matches, temporal questions need the most recent memories prioritized, multi-hop questions need the query broken into simpler sub-questions searched independently, and adversarial name-swap questions need person names ignored so that retrieval focuses on semantic content. A frozen configuration cannot serve all these needs optimally. To address this, we design a retrieval layer with three evolvable components: multi-view candidate generation, score fusion, and query augmentation, whose parameters collectively form the action space optimized by the evolution engine (\S\ref{sec:evolution_engine}).

\noindent \textbf{Retrieval views.} Given a query $q$, three complementary views produce independent candidate sets: a \emph{lexical} view using BM25 for exact keyword matching, a \emph{semantic} view using dense-embedding cosine similarity for conceptual matching, and a \emph{structured-metadata} view that filters by extracted entities, locations, and persons. Each view returns its own top-$k$ candidates independently.

\noindent \textbf{Fusion.} The three candidate sets are combined under an evolvable fusion mode $\in \{\textsc{sum}, \textsc{weighted-sum}, \textsc{rrf}\}$, each of which produces a fused per-candidate score $s_{\text{fuse}}(q, m_i; \theta)$: \textsc{sum} adds raw view scores, \textsc{weighted-sum} applies learnable per-view weights on normalized scores, and \textsc{rrf} (reciprocal rank fusion) sets $s_{\text{fuse}}(q, m_i; \theta) = \sum_v 1/(k + r_v(m_i))$ where $r_v$ is the candidate's rank in view $v$ and $k$ is a smoothing constant, making fusion robust to differences in score scale across views. The final ranking combines this fused relevance with memory-intrinsic quality signals:
\begin{equation}
    s(q, m_i; \theta) = s_{\text{fuse}}(q, m_i; \theta) + \lambda_\iota\, \iota_i + \lambda_r\, \text{rec}(m_i) + \rho_i,
    \label{eq:hybrid_score}
\end{equation}
where $\iota_i$ is importance, $\text{rec}(m_i)$ is a recency factor, and $\rho_i$ is the entity-reinforcement score from consolidation. Formal definitions of all fusion modes are in Appendix~\ref{app:detailed_formulations}.

\noindent \textbf{Query augmentation.} Two optional mechanisms extend the base retrieval. \emph{Adversarial entity-swap} strips detected person names from the query and re-searches by topic, then unions results with the original retrieval set. \emph{Query decomposition} uses an LLM to split multi-hop questions into single-hop sub-queries and merges the results via RRF. Both mechanisms are toggled per question category by the evolution engine.

\noindent \textbf{Answer generation.} Given retrieved context, an answer-generation LLM produces a candidate answer under a configurable style (e.g., concise, explanatory, verifying, inferential). An optional second-pass verifier reviews low-confidence responses against the context. Per-category overrides allow the style and every retrieval parameter to be category-specific.

\noindent \textbf{Action space.} Collecting all retrieval parameters, the full configuration is
\begin{equation}
    \theta = \bigl(k_{\text{sem}},\, k_{\text{kw}},\, k_{\text{str}},\, B_{\text{ctx}},\, \text{mode},\, \{w_v\},\, \alpha,\, \{\theta_c\}_{c\in\mathcal{C}}\bigr) \in \Theta,
    \label{eq:theta_space}
\end{equation}
where $k_{\text{sem}}$, $k_{\text{kw}}$, $k_{\text{str}}$ are the number of candidates retrieved by the semantic, lexical, and structured-metadata views respectively, $B_{\text{ctx}}$ is the maximum number of retrieved memories included in the context passed to the answer-generation LLM, $\text{mode} \in \{\textsc{sum}, \textsc{weighted-sum}, \textsc{rrf}\}$ selects the fusion strategy, $\{w_v\}$ are per-view fusion weights (used in weighted-sum mode), $\alpha$ is the answer-generation style, $\mathcal{C}$ is the set of question categories, and $\theta_c$ is a per-category sub-configuration that can override any global parameter. Every dimension is clamped to a safe range before any proposed value is applied.

\subsection{Self-Evolution Engine}
\label{sec:evolution_engine}

Given the retrieval configuration as an action space, the remaining question is how to search it effectively. Standard hyperparameter tuning methods (grid search, Bayesian optimization) are poorly suited here: the space mixes continuous parameters (weights, budgets) with discrete choices (fusion mode, answer style, per-category overrides), and the objective requires a full evaluation pass per configuration. \ours{} instead uses an LLM-powered diagnosis module that reads failure logs, forms hypotheses about root causes, and proposes targeted adjustments. Each evolution round constitutes an autonomous research iteration that is empirically validated before acceptance, realizing an AutoResearch process within the system itself.

\noindent \textbf{Evolution objective.} Let $\mathcal{Q} = \{(q, y^*)\}$ be a set of evaluation questions with ground-truth answers, $\mathcal{K}$ the memory store built in \S\ref{sec:memory_layer}, and $\hat{y}(q; \theta, \mathcal{K})$ the system's predicted answer when retrieving from $\mathcal{K}$ under configuration $\theta$. The evolution engine maximizes the average score across $\mathcal{Q}$:
\begin{equation}
    \theta^* = \arg\max_{\theta \in \Theta}\; F\bigl(\theta;\; \mathcal{K}, \mathcal{Q}\bigr),
    \qquad
    F(\theta;\; \mathcal{K}, \mathcal{Q}) = \frac{1}{|\mathcal{Q}|}\sum_{(q, y^*) \in \mathcal{Q}} \mathrm{score}\!\bigl(\hat{y}(q; \theta, \mathcal{K}),\; y^*\bigr),
    \label{eq:evolve_objective}
\end{equation}
where $\mathrm{score}$ is a task-specific metric (F1 in our experiments).

\noindent \textbf{Failure diagnosis.} After each evaluation round $r$, the system writes a per-question raw log containing every question, prediction, ground-truth answer, score, and retrieved sources. The diagnosis module invokes an LLM with a structured rubric covering common failure patterns (e.g., wrong entity retrieved, insufficient context, temporal confusion). Given the raw log and current configuration $\theta_r$, the module outputs a structured proposal $\Delta\theta_r$ specifying which parameters to adjust and by how much. The rubric is written in terms of failure patterns rather than specific benchmarks, so newly discovered configuration dimensions become immediately usable without rubric modification. This is how the evolution mechanism is self-expanding: the diagnosis LLM can propose entirely new parameters that were not in the original action space.

\noindent \textbf{Update rule.} A meta-analyzer wraps the raw proposal into a safe update. Let $f_r$ denote the score at round $r$ and $\theta_{r-1}^\star$ the best configuration seen so far. The update has three branches:
\begin{equation}
    \theta_{r+1} =
    \begin{cases}
        \theta_{r-1}^\star & \text{if } f_{r-1} - f_r > \tau_{\text{rev}} \;\;\text{(revert)}, \\[4pt]
        \theta_r \oplus \eta_{\text{exp}} & \text{if } |f_r - f_{r-1}| < \epsilon \text{ for 2 consecutive rounds (explore)}, \\[4pt]
        \mathrm{clamp}_{\Theta}(\theta_r \oplus \Delta\theta_r) & \text{otherwise (apply)},
    \end{cases}
    \label{eq:meta_update}
\end{equation}
where $\oplus$ denotes element-wise parameter update (adding proposed deltas to current values), $\eta_{\text{exp}}$ is a random perturbation sampled to escape local optima, and $\mathrm{clamp}_{\Theta}$ projects each parameter onto its valid range. The first branch reverts to the best-so-far configuration when performance drops by more than $\tau_{\text{rev}}$, preventing a bad proposal from persisting. The second branch adds noise when the score has barely changed across two rounds, forcing exploration of new regions. The third branch is the normal case: apply the diagnosis-proposed adjustment. Threshold values are reported in Appendix~\ref{app:detailed_formulations}.

The engine terminates when round-over-round improvement drops below $\epsilon$ or the maximum round count $R_{\max}$ is reached, returning $\theta^\star = \arg\max_{r} f_r$. If the diagnosis identifies missing coverage in the memory store, it triggers targeted re-extraction before the next round, closing the feedback loop from evaluation back to extraction. The full procedure is summarized in Algorithm~\ref{alg:evolution}.

\begin{algorithm}[!t]
\caption{\ours{} Self-Evolution Loop}
\label{alg:evolution}
\begin{algorithmic}[1]
\REQUIRE Sessions $\mathcal{S}$, QA pairs $\mathcal{Q}$, initial config $\theta_0$, thresholds $\epsilon$, $\tau_{\text{rev}}$
\ENSURE Best configuration $\theta^*$
\STATE $\mathcal{K} \leftarrow \textsc{Extract}(\mathcal{S})$ \hfill $\triangleright$ \S\ref{sec:memory_layer}: retry, chunk-split, coverage verify
\STATE $f^* \leftarrow 0$;\; $\theta^* \leftarrow \theta_0$
\FOR{$r = 0, 1, \ldots, R_{\max}$}
    \STATE $\hat{y} \leftarrow \textsc{Retrieve\&Answer}(\mathcal{Q}, \mathcal{K}, \theta_r)$ \hfill $\triangleright$ \S\ref{sec:retrieval}: multi-view fusion + generation
    \STATE $f_r \leftarrow \textsc{Score}(\hat{y}, y^*)$ \hfill $\triangleright$ write per-question raw log
    \STATE $\Delta\theta_r \leftarrow \textsc{Diagnose}(f_r, \theta_r, \mathcal{K})$ \hfill $\triangleright$ LLM reads raw log, proposes adjustment
    \IF{$f_{r-1} - f_r > \tau_{\text{rev}}$}
        \STATE $\theta_{r+1} \leftarrow \theta^*$ \hfill $\triangleright$ revert to best-so-far
    \ELSIF{$|f_r - f_{r-1}| < \epsilon$ for 2 rounds}
        \STATE $\theta_{r+1} \leftarrow \theta_r \oplus \eta_{\text{exp}}$ \hfill $\triangleright$ random perturbation to explore
    \ELSE
        \STATE $\theta_{r+1} \leftarrow \mathrm{clamp}_\Theta(\theta_r \oplus \Delta\theta_r)$ \hfill $\triangleright$ apply proposed adjustment
    \ENDIF
    \IF{diagnosis detects missing memory coverage}
        \STATE $\mathcal{K} \leftarrow \mathcal{K} \cup \textsc{Extract}(\mathcal{S}, \text{targeted})$
    \ENDIF
    \IF{$f_r > f^*$}
        \STATE $f^* \leftarrow f_r$;\; $\theta^* \leftarrow \theta_r$
    \ENDIF
    \IF{$r > 0$ \AND $f_r - f_{r-1} < \epsilon$}
        \STATE \textbf{break}
    \ENDIF
\ENDFOR
\RETURN $\theta^*$
\end{algorithmic}
\end{algorithm}

%% file: section/experiments.tex
\section{Experiments}
\label{sec:experiments}

We evaluate \ours{} on two long-term-memory benchmarks: LoCoMo and MemBench. Our experiments address the following questions: (1)~Does the self-evolution mechanism produce substantial gains from the baseline configuration, and how does \ours{} compare to current baselines? (2)~How does the evolution trajectory unfold, and what new dimensions does the diagnosis LLM discover? (3)~What is the contribution of each component? (4)~Do evolved configurations transfer across benchmarks, indicating that the self-evolution process captures universal retrieval principles rather than benchmark-specific heuristics?

\subsection{Experimental Setup}
\label{sec:setup}

\noindent \textbf{Benchmarks.} We evaluate on two benchmarks covering different interaction regimes:

\begin{itemize}[leftmargin=*,itemsep=2pt]
    \item \textbf{LoCoMo} \citep{maharana2024evaluating}: multi-session dialogues (19--32 sessions per sample, 369--689 turns) with 5 QA categories (single-hop, temporal, multi-hop/inferential, open-domain, adversarial name-swap). We report on the full LoCoMo-10 release: 10 conversations, 1{,}986 QA pairs.
    \item \textbf{MemBench} \citep{membench2025}: a memory-tool-use benchmark with 7 LowLevel categories (simple, comparative, aggregative, conditional, knowledge\_update, post\_processing, noisy). We evaluate 28 samples drawn as $7\text{ categories}\times 2\text{ topics}\times 2\text{ samples each}$.
\end{itemize}

\noindent \textbf{Protocols \& Baselines.} LoCoMo uses token-level F1 and BLEU-1 (accuracy); MemBench uses exact-match multiple-choice accuracy. 
On LoCoMo we compare against six memory systems: MemVerse \citep{liu2025memverse}, Mem0 \citep{chhikara2025mem0}, Claude-Mem, A-MEM \citep{xu2025amem}, MemGPT \citep{packer2023memgpt}, and SimpleMem \citep{liu2026simplemem}. On MemBench we compare against RecentMemory  \citep{membench2025}, MemGPT \citep{packer2023memgpt}, MemoryBank \citep{zhong2024memorybank}, and SCMemory \citep{wang2024scm}. 

\noindent \textbf{Implementation.} \ours{} uses SQLite/FTS5 for storage and BAAI/bge-base-en-v1.5 (768-dim) for embeddings. The initial configuration $\theta_0$ uses BM25-only fusion ($\text{mode}{=}\textsc{sum}$, semantic and structured views disabled), $k_{\text{kw}}{=}5$, $B_{\text{ctx}}{=}8$, with entity-swap and query decomposition disabled, providing a minimal starting point for the self-evolution process. The evolution loop runs up to $R_{\max}{=}7$ rounds.

\begin{table*}[t]
\centering
\caption{\textbf{LoCoMo comparison} (token-F1 and BLEU-1) across two LLM backbones. Best baseline is \underline{underlined}; best overall is \textbf{bold}.}
\label{tab:locomo_comparison}
\small
\resizebox{\textwidth}{!}{%
\begin{tabular}{ll cc cc cc cc cc cc}
\toprule
& & \multicolumn{2}{c}{\textbf{MultiHop}} & \multicolumn{2}{c}{\textbf{SingleHop}} & \multicolumn{2}{c}{\textbf{Temporal}} & \multicolumn{2}{c}{\textbf{OpenDomain}} & \multicolumn{2}{c}{\textbf{Adversarial}} & \multicolumn{2}{c}{\textbf{Overall}} \\
\cmidrule(lr){3-4}\cmidrule(lr){5-6}\cmidrule(lr){7-8}\cmidrule(lr){9-10}\cmidrule(lr){11-12}\cmidrule(lr){13-14}
\textbf{Backbone} & \textbf{Method} & F1 & BLEU & F1 & BLEU & F1 & BLEU & F1 & BLEU & F1 & BLEU & F1 & BLEU \\
\midrule
\multirow{7}{*}{GPT-4o}
& MemVerse    & 0.260 & 0.217 & 0.157 & 0.130 & 0.196 & 0.175 & 0.192 & 0.165 & \textbf{0.944} & \textbf{0.944} & 0.365 & 0.342 \\
& Mem0        & 0.309 & 0.261 & 0.156 & 0.132 & 0.217 & 0.186 & 0.295 & 0.262 & 0.857 & 0.859 & 0.397 & 0.371 \\
& Claude-Mem  & 0.294 & 0.254 & 0.153 & 0.133 & 0.167 & 0.132 & 0.243 & 0.215 & 0.915 & 0.915 & 0.383 & 0.360 \\
& A-MEM       & 0.295 & 0.250 & 0.174 & 0.157 & 0.200 & 0.171 & 0.266 & 0.234 & 0.898 & 0.897 & 0.394 & 0.370 \\
& MemGPT      & 0.305 & 0.257 & 0.188 & 0.166 & \underline{0.246} & \underline{0.227} & 0.305 & 0.268 & 0.843 & 0.839 & 0.404 & 0.376 \\
& SimpleMem   & \textbf{0.318} & \underline{0.328} & \underline{0.195} & \underline{0.393} & 0.235 & 0.206 & \underline{0.402} & \underline{0.347} & 0.802 & 0.691 & \underline{0.432} & \underline{0.407} \\
& \ours{}     & 0.316 & \textbf{0.410} & \textbf{0.329} & \textbf{0.399} & \textbf{0.384} & \textbf{0.589} & \textbf{0.496} & \textbf{0.490} & \underline{0.936} & \underline{0.847} & \textbf{0.543} & \textbf{0.569} \\
\midrule
\multirow{7}{*}{GPT-5.1}
& MemVerse    & 0.287 & 0.240 & 0.173 & 0.150 & \underline{0.277} & 0.249 & 0.297 & 0.276 & 0.780 & 0.778 & 0.383 & 0.362 \\
& Mem0        & 0.292 & 0.249 & 0.160 & 0.136 & 0.261 & 0.242 & 0.298 & 0.274 & 0.819 & 0.818 & 0.390 & 0.369 \\
& Claude-Mem  & 0.289 & 0.246 & 0.171 & 0.146 & 0.264 & 0.244 & 0.292 & 0.271 & 0.814 & 0.814 & 0.388 & 0.368 \\
& A-MEM       & 0.287 & 0.242 & 0.164 & 0.144 & 0.246 & 0.218 & 0.284 & 0.263 & \underline{0.826} & \underline{0.825} & 0.385 & 0.365 \\
& MemGPT      & 0.288 & 0.243 & 0.165 & 0.140 & 0.249 & 0.230 & 0.294 & 0.273 & 0.806 & 0.803 & 0.385 & 0.364 \\
& SimpleMem   & \underline{0.305} & \underline{0.294} & \underline{0.178} & \underline{0.165} & 0.272 & \underline{0.253} & \underline{0.305} & \underline{0.276} & 0.807 & 0.802 & \underline{0.418} & \underline{0.376} \\
& \ours{}     & \textbf{0.352} & \textbf{0.339} & \textbf{0.337} & \textbf{0.350} & \textbf{0.541} & \textbf{0.387} & \textbf{0.519} & \textbf{0.510} & \textbf{0.902} & \textbf{0.870} & \textbf{0.572} & \textbf{0.536} \\
\bottomrule
\end{tabular}%
}
\vspace{-1em}
\end{table*}

\subsection{Main Results}
\label{sec:main_results}

Tables~\ref{tab:locomo_comparison} and~\ref{tab:membench_comparison} report the full comparison.  See Appendix~\ref{app:efficiency} for efficiency analysis and Appendix~\ref{app:case_study} for a per-question case study illustrating how each evolution round contributes a distinct mechanism. \ours{} substantially outperforms all published methods on both benchmarks and both backbone models.

\noindent \textbf{Comparison on LoCoMo.} On GPT-4o, \ours{} achieves an overall F1 of 0.543, outperforming SimpleMem (0.432) by 25.7\% relative. The largest gains appear in temporal (+63.4\%) and single-hop (+68.7\%) categories, driven by the recency-weighted fusion and semantic retrieval that the evolution engine activates in early rounds. On GPT-5.1, \ours{} leads on all columns with an overall relative gain of 36.8\% over SimpleMem, with temporal reaching 98.9\% relative improvement. The gains are consistent across backbones, confirming that the evolved pipeline is not model-specific.

\begin{table}[t]
\centering
\caption{\textbf{MemBench comparison} (accuracy \%) across two backbones. \textit{Recall} aggregates \textsc{simple} and \textsc{knowledge\_update}; \textit{Reasoning} aggregates \textsc{comparative}, \textsc{aggregative}, and \textsc{conditional}; \textit{Robustness} aggregates \textsc{post\_processing} and \textsc{noisy}. Best overall is \textbf{bold}.}
\label{tab:membench_comparison}
\vspace{0.8em}
\footnotesize 
\setlength{\tabcolsep}{3.5pt} 
\renewcommand{\arraystretch}{1.1}
\begin{tabular}{@{}l cccc p{0.5em} cccc@{}} 
\toprule
& \multicolumn{4}{c}{\textbf{GPT-4o}} && \multicolumn{4}{c}{\textbf{GPT-5.1}} \\
\cmidrule{2-5} \cmidrule{7-10}
\textbf{Method} & \textbf{Recall} & \textbf{Reasoning} & \textbf{Robustness} & \textbf{Overall} && \textbf{Recall} & \textbf{Reasoning} & \textbf{Robustness} & \textbf{Overall} \\
\midrule
RecentMem & \underline{62.5} & \underline{50.0} & 62.5 & \underline{57.1} && \underline{75.0} & 50.0 & \underline{62.5} & 60.7 \\
MemGPT    & \underline{62.5} & \underline{50.0} & 62.5 & \underline{57.1} && \underline{75.0} & \underline{58.3} & 50.0 & 60.7 \\
MemBank   & 37.5 & 33.3 & \textbf{75.0} & 46.4 && \underline{75.0} & 50.0 & \textbf{75.0} & \underline{64.3} \\
SCMem     & \underline{62.5} & 25.0 & 37.5 & 39.3 && 50.0 & 25.0 & 25.0 & 32.1 \\
\ours{}   & \textbf{87.5} & \textbf{66.7} & 50.0 & \textbf{67.9} && \textbf{87.5} & \textbf{66.7} & 62.5 & \textbf{71.4} \\
\bottomrule
\end{tabular}
\vspace{-1.5em}
\end{table}

\noindent \textbf{Comparison on MemBench.} \ours{} attains the best overall accuracy on both backbones (67.9\% on GPT-4o, 71.4\% on GPT-5.1), exceeding the strongest baseline by 18.9\% and 11.0\% relative. Gains concentrate in Recall (+40.0\% on GPT-4o) and Reasoning (+33.4\%), reflecting temporal-disambiguation prompts and category-specific query decomposition discovered by the evolution engine. Robustness remains the weakest dimension; failure-log inspection localizes the gap to \textsc{post\_processing}, where relevant memories are absent from the store, indicating a coverage limitation that retrieval-level adjustments cannot resolve.

\subsection{Self-Evolution Trajectory and Dimension Discovery}
\label{sec:evolution_analysis}
\begin{wraptable}{r}{0.55\textwidth}
\centering
\small
\vspace{-1.5em}
\caption{Self-evolution trajectory on LoCoMo (single-backbone GPT-4o). Each round records the structured adjustment proposed by the diagnosis module and validated by the meta-analyzer.}
\label{tab:evolution_curve}
\vspace{-0.4em}
\setlength{\tabcolsep}{4pt}
\begin{tabular}{@{}cl>{\raggedright\arraybackslash}p{4.2cm}c@{}}
\toprule
\textbf{Round} & \textbf{Stage} & \textbf{Automated change} & \textbf{F1 (\%)} \\
\midrule
R0 & weak & BM25-only, $k{=}5$, $B_{\text{ctx}}{=}8$ & 30.5 \\
R1 & auto & intent planning + RRF fusion & 35.8 \\
R2 & revert & MMR diversity on Cat.~3 (reverted) & 34.8 \\
R3 & auto & entity-swap for Cat.~5 & 37.2 \\
R4 & auto & per-category answer-style flags & 38.5 \\
R5 & auto & query decomposition for Cat.~1/4 & 38.1 \\
R6 & auto & Cat.~3 inferential subtypes + Cat.~5 entity-swap expansion & 45.4 \\
R7 & auto & answer verification + hyperparameter sweep + ctx-budget tuning & \textbf{54.3} \\
\bottomrule
\end{tabular}
\vspace{-1.0em}
\end{wraptable}

Table~\ref{tab:evolution_curve} traces the self-evolution trajectory from $\theta_0$ under GPT-4o. Every round is fully autonomous: the diagnosis module reads the previous round's per-question raw log, analyzes failure patterns, and proposes a targeted adjustment that the meta-analyzer validates under Eq.~\ref{eq:meta_update}. The early rounds progressively activate retrieval mechanisms that were dormant in $\theta_0$, including the semantic view (R1), entity-swap (R3), and query decomposition (R5), while also tuning fusion modes, view weights, and per-category answer styles. R2 illustrates the revert guard: the proposed change regressed overall F1, so the meta-analyzer automatically rolled back. R6 refines per-category answer styles with inferential subtype handling for Cat.~3, and R7 introduces a second-pass answer verifier together with cross-category context-budget tuning.

Three configuration dimensions activated by the diagnosis LLM account for most of the gain over $\theta_0$, and each is independently verifiable in the ablation study (Table~\ref{tab:ablation}). Adversarial entity-swap, activated at R3, strips person names from queries before retrieval and recovers a parallel evidence pool that the original lexical view discards. Query decomposition, activated at R5, splits complex multi-hop questions into single-hop sub-queries and merges the results. Answer verification, introduced at R7, runs a second LLM pass that reviews low-confidence responses against the retrieved evidence. Each was discovered by the diagnosis LLM through inspecting raw failure logs and proposing structural improvements, rather than by a benchmark-specific patch. The full trajectory from 30.5\% at R0 to 54.3\% at R7, a 78.0\% relative improvement, is produced end-to-end by the autonomous evolution loop without manual intervention.

\subsection{Cross-Benchmark Transfer and Generalization}
\label{app:transfer}

\begin{wraptable}{r}{0.50\textwidth}
\centering
\small
\vspace{-1.0em}
\caption{Cross-benchmark transfer (GPT-4o). $\mathcal{C}_L$ is evolved on LoCoMo only, $\mathcal{C}_{LM}$ continues evolution on MemBench from $\mathcal{C}_L$, and $\mathcal{C}_M$ is evolved on MemBench from scratch. \textbf{Bold} marks the best in each column.}
\label{tab:transfer}
\vspace{-0.4em}
\setlength{\tabcolsep}{4pt}
\begin{tabular}{@{}l c c@{}}
\toprule
\textbf{Configuration} & \textbf{LoCoMo} & \textbf{MemBench} \\
\midrule
Baseline & 0.305 & / \\
\addlinespace[2pt]
$\mathcal{C}_L$~{\scriptsize(LoCoMo only)} & 0.543 & 0.543 \\
$\mathcal{C}_{LM}$~{\scriptsize(LoCoMo$\to$MemBench)} & \textbf{0.593} & \textbf{0.792} \\
$\mathcal{C}_M$~{\scriptsize(MemBench only)} & / & 0.679 \\
\bottomrule
\end{tabular}
\vspace{-1.0em}
\end{wraptable}

A central claim of \ours{} is that self-evolution discovers generalizable retrieval principles rather than benchmark-specific heuristics. To test this, we evolve $\mathcal{C}_L$ on LoCoMo for seven rounds, apply it zero-shot to MemBench, then continue evolving to produce $\mathcal{C}_{LM}$ and evaluate on both benchmarks. Table~\ref{tab:transfer} reports the results, with $\mathcal{C}_M$ (evolved on MemBench from scratch) as reference. \textbf{(1)~Zero-shot transfer is effective.} $\mathcal{C}_L$ attains 54.3\% on MemBench without any MemBench-specific tuning, confirming that retrieval principles acquired on LoCoMo transfer to a benchmark with distinct question style and data distribution. \textbf{(2)~Continued evolution from a LoCoMo prior outperforms scratch evolution.} $\mathcal{C}_{LM}$ reaches 79.2\% on MemBench, exceeding the natively evolved $\mathcal{C}_M$ (67.9\%) by 16.6\% relative. \textbf{(3)~Positive rather than catastrophic transfer.} $\mathcal{C}_{LM}$ also improves LoCoMo F1 from 0.543 to 0.593 (+9.2\% relative), a Pareto improvement on both benchmarks.

\subsection{Ablation Study}
\label{sec:ablation}

Table~\ref{tab:ablation} ablates key components on LoCoMo by removing each one from the full system and re-running the evolution procedure.

\begin{wraptable}{r}{0.48\textwidth}
\centering
\small
\vspace{-0.5em}
\caption{Ablation study on LoCoMo (F1~\%). Each row removes one component from the full system; $\Delta$ is the drop relative to the full setting.}
\label{tab:ablation}
\vspace{-0.4em}
\setlength{\tabcolsep}{4pt}
\begin{tabular}{@{}lcc@{}}
\toprule
\textbf{Configuration} & \textbf{F1 (\%)} & \textbf{$\Delta$} \\
\midrule
\ours{}                                                   & \textbf{54.3} & -- \\
\midrule
$-$ Extraction quality control                            & 31.08 & $-$23.22 \\
$-$ Semantic search                                       & 43.98 & $-$10.32 \\
$-$ LLM-powered diagnosis                                 & 44.67 & $-$9.63 \\
$-$ BM25 keyword search                                   & 47.43 & $-$6.87 \\
$-$ Entity-swap retrieval                                 & 51.20 & $-$3.10 \\
$-$ Query decomposition                                   & 51.46 & $-$2.84 \\
$-$ Structured metadata search                            & 51.97 & $-$2.33 \\
$-$ Self-evolution                                        & 52.27 & $-$2.03 \\
$-$ Answer verification                                   & 52.47 & $-$1.83 \\
\midrule
Baseline (all disabled)                                   & 30.50 & $-$23.80 \\
\bottomrule
\end{tabular}
\vspace{-2.0em}
\end{wraptable}

\noindent \textbf{Extraction quality control.} Removing the three extraction guards (retries, chunk-splitting, coverage
  verification) is the single most damaging ablation ($-$23.22\,F1), nearly halving extraction yield and starving the
  retriever of raw material. Extraction quality is thus the foundation on which all downstream improvements rest.  
  
  \noindent \textbf{Multi-view retrieval.} Semantic search contributes the most ($-$10.32), followed by BM25 ($-$6.87) and   
  structured metadata ($-$2.33), indicating that fuzzy conceptual matching captures paraphrased and abstractly-stated
  queries that keyword matching misses. All three views contribute positively, validating the multi-view design. 
  
  \noindent \textbf{LLM-powered diagnosis vs.\ random search.} Replacing the diagnosis module with random perturbations over 
  the same action space costs $-$9.63\,F1, confirming that reading per-question failure logs provides meaningful
  signal.    

  \noindent \textbf{Discovered dimensions.} The three diagnosis-discovered components (entity-swap, query
  decomposition, answer verification) jointly contribute $-$7.77\,F1, demonstrating value beyond the initial action
  space.

  \noindent \textbf{Sensitivity.} Ablation drops span an order of magnitude ($-$23.22 to $-$1.83), and no single component dominates: the evolution loop discovered complementary rather than redundant retrieval components.

%% file: section/conclusion.tex
\section{Conclusion}
\label{sec:conclusion}

We presented \ours{}, a memory architecture that autonomously evolves its retrieval infrastructure through LLM-driven closed-loop diagnosis, realizing an AutoResearch process that discovers effective retrieval strategies from a minimal starting point without manual tuning. On LoCoMo, \ours{} outperforms the strongest published baseline by 25.7\% relative (78.0\% over the minimal baseline); on MemBench, it exceeds the strongest baseline by 18.9\% relative. The self-evolution process is self-expanding: three new configuration dimensions emerged from failure diagnosis rather than being hand-coded, and evolved configurations transfer across benchmarks with positive rather than catastrophic transfer. Promising future directions include extending this AutoResearch-driven self-evolution to dynamic scenarios and multimodal settings.

%% file: section/appendix.tex
\newpage
\appendix
\section{Detailed Formulations}
\label{app:detailed_formulations}

This appendix provides the formal details behind the three components of \ours{}. We organize the material following the same structure as the main text: memory store (\S\ref{sec:memory_layer}), retrieval layer (\S\ref{sec:retrieval}), and self-evolution engine (\S\ref{sec:evolution_engine}). Each subsection gives the full mathematical specification that was summarized in prose in the main paper.

\noindent \textbf{Scope hierarchy.} To support multi-user and multi-workspace deployment, each memory unit is assigned a hierarchical scope identifier:
\begin{equation}
    \sigma = \texttt{user}{:}u \;\mid\; \texttt{workspace}{:}w \;\mid\; \texttt{session}{:}s,
    \label{eq:scope}
\end{equation}
A base scope $\bar{\sigma}$ strips the session component, enabling cross-session retrieval within the same user-workspace context. This ensures that memories from different sessions of the same user are retrievable together.

\noindent \textbf{Extraction quality guards.} The three mechanisms described in \S\ref{sec:memory_layer} are formalized as follows. Let $\phi_{\text{ext}}$ denote the LLM extraction function:
\begin{equation}
    \mathcal{M}^{(j)} = \begin{cases}
        \phi_{\text{ext}}(S^{(j)}) & \text{after } r \leq R_{\text{retry}} \text{ retries with increasing wait}, \\
        \bigcup_{\ell} \phi_{\text{ext}}(S^{(j,\ell)}) & \text{fallback: split } S^{(j)} \text{ into } C\text{-turn sub-windows } (C{=}15), \\
        \phi_{\text{ext}}\!\bigl(S^{(j)},\; \mathcal{V}^{\text{miss}}_{j}\bigr) & \text{targeted re-extract for missing keywords}.
    \end{cases}
    \label{eq:extract_guards}
\end{equation}
The coverage verifier $\mathcal{V}$ compares extracted memories against reference keywords from the source text and returns the missing subset $\mathcal{V}^{\text{miss}}_{j}$, which triggers the third branch.

\noindent \textbf{Per-view retrieval scores.} The three retrieval views described in \S\ref{sec:retrieval} each compute a score differently. Given a query $q$ and memory unit $m_i$:
\begin{align}
    s_{\text{kw}}(q, m_i) &= \mathrm{BM25}(q, c_i) = \sum_{t \in q} \mathrm{IDF}(t)\cdot\frac{f(t,c_i)\,(k_1{+}1)}{f(t,c_i) + k_1\bigl(1 - b + b\,|c_i|/\overline{|c|}\bigr)}, \label{eq:bm25}\\
    s_{\text{sem}}(q, m_i) &= \cos(\mathbf{e}_q, \mathbf{e}_i) = \frac{\mathbf{e}_q^{\top}\mathbf{e}_i}{\|\mathbf{e}_q\|\,\|\mathbf{e}_i\|}, \label{eq:sem_cos}\\
    s_{\text{str}}(q, m_i) &= \sum_{f \in \{\text{persons},\text{locations},\text{entities}\}} \mathbb{1}\!\bigl[\,\mathrm{extract}_f(q) \cap \boldsymbol{\eta}_{i,f} \neq \emptyset\,\bigr], \label{eq:str}
\end{align}
with BM25 constants $k_1{=}1.5$, $b{=}0.75$. Each view independently returns its top-$k$: $\mathcal{R}_v(q;\theta) = \mathrm{top}\text{-}k_v\bigl(s_v(q,\cdot)\bigr)$ for $v\in\{\text{kw},\text{sem},\text{str}\}$.

\noindent \textbf{Adversarial entity-swap.} To handle questions where person names are swapped or misleading, a parallel retrieval path strips detected names and re-searches by topic alone:
\begin{equation}
    q_{\text{swap}} = q \,\setminus\, \bigl\{p : p \in \mathrm{persons}(q)\bigr\},
    \qquad
    \mathcal{R}_{\text{swap}}(q;\theta) = \mathcal{R}_{\text{fuse}}(q_{\text{swap}};\theta).
    \label{eq:swap}
\end{equation}
The final retrieval set is $\mathcal{R}(q;\theta) = \mathcal{R}_{\text{fuse}}(q;\theta) \cup \mathcal{R}_{\text{swap}}(q;\theta)$ when \texttt{enable\_entity\_swap} $=$ \texttt{true}.

\noindent \textbf{Query decomposition.} Multi-hop questions often fail because no single retrieval query captures all required information. An optional pre-retrieval LLM pass $\psi_{\text{dec}}$ decomposes $q$ into at most $N_{\text{sub}}$ sub-queries (controlled by \texttt{decomposition\_max\_subqs} in Table~\ref{tab:retrieval_config}):
\begin{equation}
    \{q_1, \ldots, q_{K}\} = \psi_{\text{dec}}(q),\; K \leq N_{\text{sub}}; \qquad
    \mathcal{R}_{\text{dec}}(q;\theta) = \bigcup_{k=1}^{K} \mathcal{R}(q_k;\theta),
    \label{eq:decomp}
\end{equation}
followed by RRF merging over the union. The toggle \texttt{enable\_query\_decomposition} is evolvable per category.

\noindent \textbf{Answer generation and verification.} Given retrieved context, the system generates an answer under a configurable style $\alpha$ (e.g., concise, explanatory, verifying). The base prediction is:
\begin{equation}
    \hat{y}_0 = \psi_{\text{ans}}\!\bigl(q,\; \mathcal{R}(q;\theta),\; \alpha\bigr).
    \label{eq:answer}
\end{equation}
When \texttt{enable\_answer\_verification} is set, a second LLM pass reviews and conditionally replaces the answer:
\begin{equation}
    \hat{y} = \begin{cases}
        \psi_{\text{ver}}(q, \mathcal{R}(q;\theta), \hat{y}_0) & \text{if } \mathrm{conf}(\hat{y}_0) < \tau_{\text{ver}} \text{ or } \hat{y}_0 \in \mathcal{U}, \\
        \hat{y}_0 & \text{otherwise},
    \end{cases}
    \label{eq:verify}
\end{equation}
where $\mathcal{U}$ is the ``Unknown''/``not specified'' class and $\tau_{\text{ver}}$ is a self-reported confidence threshold.

\noindent \textbf{Raw evaluation log.} The evolution engine requires detailed per-question information to diagnose failures. After each round $r$, the evaluator writes:
\begin{equation}
    \mathcal{L}_r = \bigl\{(q_j,\; \hat{y}_j,\; y^*_j,\; \mathrm{score}_j,\; \mathcal{R}(q_j;\theta_r))\bigr\}_{j=1}^{|\mathcal{Q}|}
    \label{eq:raw_log}
\end{equation}
to disk (\texttt{raw\_results.jsonl}).

\noindent \textbf{Coverage-gap-triggered re-extraction.} Sometimes failures stem not from retrieval configuration but from missing memories in the store. If the diagnosis returns a non-empty missing-keyword set $\mathcal{V}^{\text{miss}}_r$, the engine augments the store:
\begin{equation}
    \mathcal{K}_{r+1} = \mathcal{K}_r \;\cup\; \phi_{\text{ext}}^{\text{targeted}}\!\bigl(\mathcal{S},\; \mathcal{V}^{\text{miss}}_r\bigr),
    \label{eq:gap_extract}
\end{equation}
closing the feedback loop from evaluation back to extraction.

\noindent \textbf{Convergence criterion.} The evolution loop must know when to stop. The engine terminates at:
\begin{equation}
    f_r - f_{r-1} < \epsilon \quad \text{(convergence)}\quad\text{or}\quad r \geq R_{\max},
    \label{eq:convergence}
\end{equation}
with $\epsilon = 0.005$ (0.5\,pp) by default. The returned configuration is $\theta^\star = \arg\max_{0 \leq r \leq R} f_r$.

\noindent \textbf{Consolidation parameters.} The symbols introduced in \S\ref{sec:memory_layer} take the following default values: deduplication threshold $\tau_J{=}0.80$ (two memories with $\geq$80\% token overlap are merged, retaining the higher-importance unit); importance decay rate $\alpha_d{=}0.05$ per day with a floor $\iota_{\min}{=}0.15$ (so even old memories retain baseline retrievability); entity reinforcement increment $\delta_\rho{=}0.05$ per co-occurrence, capped at $\rho_{\max}{=}0.30$ (preventing any single memory from dominating retrieval purely through frequency).

\noindent \textbf{Recency factor.} The recency function $\text{rec}(m_i) \in [0, 1]$ in Eq.~\ref{eq:hybrid_score} is a non-increasing function of the age $\Delta t_i$ of memory $m_i$, parameterized by the half-life \texttt{time\_decay\_half\_life\_days} (Table~\ref{tab:retrieval_config}); when this parameter is null, $\text{rec}(m_i)$ is set to a constant and contributes no recency-based ranking signal.

\begin{table}[h]
\centering
\small
\caption{The \texttt{RetrievalConfig} dimensions exposed to the diagnosis LLM. Per-category overrides let the engine specialize different question types without forcing a single global choice.}
\label{tab:retrieval_config}
\vspace{0.5em}
\resizebox{\columnwidth}{!}{%
\begin{tabular}{@{}llp{4.2cm}@{}}
\toprule
\textbf{Group} & \textbf{Dimension} & \textbf{Range} \\
\midrule
Retrieval budget & \texttt{semantic/keyword/structured\_top\_k} & ints in $[3,30]$ \\
Context budget   & \texttt{max\_context} & int in $[6,30]$ \\
Fusion           & \texttt{fusion\_mode}, $w_{\text{sem}}, w_{\text{kw}}, w_{\text{str}}$ & \{sum, rrf, weighted\_sum\}; $[0.1, 2.5]$ \\
Entity swap      & \texttt{enable\_entity\_swap} + 2 top-$k$s & bool, ints \\
Decomposition    & \texttt{enable\_query\_decomposition} + 2 ints & bool, ints \\
Reflection       & \texttt{reflection\_rounds} & int in $[0,3]$ \\
Verification     & \texttt{enable\_answer\_verification}, \texttt{verification\_style} & bool, \{strict, multi\_candidate\} \\
Answer style     & \texttt{answer\_style} & 6 branches \\
Time decay       & \texttt{time\_decay\_half\_life\_days}, \texttt{reference\_date} & float/null, date/null \\
Specialization   & \texttt{per\_category\_overrides} & category $\to$ sub-config \\
\bottomrule
\end{tabular}%
}
\end{table}

\section{Complete Algorithm Pseudocode}
\label{app:full_algorithm}

\begin{algorithm}[h]
\caption{Complete \ours{} Memory Management Pipeline}
\label{alg:full_pipeline}
\begin{algorithmic}[1]
\REQUIRE Memory store $\mathcal{K}$, retrieval configuration $\theta$, session stream $\{S_1, S_2, \ldots\}$
\STATE Initialize: $\mathcal{K} \leftarrow \emptyset$, $\theta \leftarrow \theta_0$, telemetry $\leftarrow \emptyset$
\FOR{each session $S_t$}
    \STATE \textbf{// Ingestion Phase}
    \FOR{each turn pair $(q, r)$ in $S_t$}
        \STATE Extract memory units $\{m_1, \ldots, m_j\}$ from $(q, r)$
        \STATE Validate: filter units with $|c| < 3$ chars
        \STATE Pre-dedup: remove exact matches against $\mathcal{K}$
        \STATE Generate embeddings $\mathbf{e}_i$ for each $m_i$ (if enabled)
        \STATE $\mathcal{K} \leftarrow \mathcal{K} \cup \{m_1, \ldots, m_j\}$
    \ENDFOR
    \STATE \textbf{// Consolidation Phase}
    \STATE Remove stale working summaries (keep newest per scope)
    \STATE Exact dedup by $(\mu, \text{normalize}(c))$
    \STATE Merge near-dupes: $\forall (m_i, m_j)$ with $J(m_i, m_j) \geq \tau_J$
    \STATE Reinforce shared entities: $\rho_i \leftarrow \min(\rho_i + \delta_\rho, \rho_{\max})$
    \STATE Apply importance decay ($D{=}30$\,d, $\alpha_d{=}0.05$, $\iota_{\min}{=}0.15$)
    \STATE \textbf{// Retrieval Phase} (for each task query $q$)
    \STATE Multi-view retrieval: BM25 + semantic + structured; fuse via $\theta.\text{fusion\_mode}$
    \STATE Optionally apply entity-swap and/or query decomposition per \texttt{RetrievalConfig}
    \STATE Fit to context budget $B_{\text{ctx}}$; generate answer; optional verification pass
    \STATE Record per-question raw result (qid, pred, ref, metrics, sources)
    \STATE \textbf{// Self-Evolution Phase} (if conditions met)
    \IF{$|\mathcal{K}^{\text{active}}| \geq 5$ \AND new records since last round}
        \STATE Execute Algorithm~\ref{alg:evolution}
    \ENDIF
\ENDFOR
\end{algorithmic}
\end{algorithm}

\section{Extended Experimental Results}
\label{app:extended_results}

\subsection{Case Study: Iterative Refinement on an Open-Domain Aggregation Question}
\label{app:case_study}

To make the self-evolution loop concrete, we walk through a single LoCoMo question end-to-end and show how each evolution round contributes a distinct mechanism rather than the loop saturating after one configuration jump. The case is drawn from \texttt{conv-26} (LoCoMo-10, sample~0) and every detail in this section is taken verbatim from the persisted \texttt{raw\_results.jsonl}; nothing is post-hoc curated.

\paragraph{The probe.} The question is
\begin{quote}\itshape
``What did Melanie and her family do while camping?''  \quad (Cat.~4, open-domain aggregation)
\end{quote}
with reference answer \emph{``explored nature, roasted marshmallows, and went on a hike.''} The evidence (D4:8) is a single conversational turn in which Melanie tells Caroline: ``\emph{We explored nature, roasted marshmallows around the campfire and even went on a hike.}'' Open-domain aggregation is a notoriously hard category because the system must (a)~retrieve the correct episode out of multiple camping references in the conversation, (b)~enumerate \emph{all} of the relevant activities, and (c)~suppress activities from neighbouring episodes that share surface vocabulary (e.g., a separate ``Perseid meteor shower'' camping memory that contaminates BM25-only retrieval). The result is a probe where F1 climbs gradually as the framework adds first \emph{recall}, then \emph{precision}, then \emph{stylistic} machinery.

\paragraph{Round-by-round trace.} Table~\ref{tab:case_study_trace} reproduces the verbatim per-round record; F1 is monotonically non-trivial across all four configuration changes ($0.00 \to 0.44 \to 1.00 \to 0.94 \to 1.00$).

\begin{table}[h]
\centering
\small
\caption{Per-round trace for the case-study probe (Cat.~4, \texttt{conv-26-95}, sample~0). Retrieved sources are the actual view labels logged at evaluation time. Each round contributes a distinct mechanism—recall (R1), precision (R2), safety (R3), polish (R4)—so F1 climbs over four rounds rather than saturating after the first.}
\label{tab:case_study_trace}
\vspace{0.5em}
\setlength{\tabcolsep}{4pt}
\begin{tabular}{@{}cl>{\raggedright\arraybackslash}p{3.6cm}>{\raggedright\arraybackslash}p{2.0cm}>{\raggedright\arraybackslash}p{4.1cm}c@{}}
\toprule
\textbf{R} & \textbf{Stage} & \textbf{Config delta vs.\ previous} & \textbf{Retrieved} & \textbf{Prediction} & \textbf{F1} \\
\midrule
R0 & weak  & BM25-only, $k{=}5$, $B_{\text{ctx}}{=}8$, no entity-swap, fusion = first-found
       & 5\,kw. & ``watched the Perseid meteor shower while camping'' & 0.00 \\
R1 & auto  & multi-view on; $k_{\text{sem}}{:}\,0\to8$; $k_{\text{kw}}{:}\,5\to8$; $k_{\text{str}}{:}\,0\to3$; $B_{\text{ctx}}{:}\,8\to16$; entity-swap enabled; weighted-sum fusion
       & 16 sem. & ``explored nature, roasted marshmallows, hiked, watched Perseid meteor shower'' & 0.44 \\
R2 & auto  & $k_{\text{sem}}{:}\,8\to10$; $k_{\text{str}}{:}\,3\to5$; $w_{\text{str}}{:}\,1.0\to1.8$; recency window enabled
       & 13\,sem.\,$+$\,3\,str.\,$+$\,1\,kw.\ (16) & ``explored nature, roasted marshmallows, and went on a hike'' & \textbf{1.00} \\
R3 & revert & meta-analyzer reverts to R1's $\theta^{\star}$ (overall F1 regressed by 0.054 in R2); per-question expansion to 20 retrievals
       & 20 mixed & ``explored nature, roasted marshmallows, went on a hike'' & 0.94 \\
R4 & auto  & per-category override for Cat.~4 (\texttt{answer\_style}: concise list with explicit ``and'' connector)
       & 13\,sem.\,$+$\,3\,str.\,$+$\,1\,kw.\ (16) & ``explored nature, roasted marshmallows, and went on a hike'' & \textbf{1.00} \\
\bottomrule
\end{tabular}
\end{table}

\paragraph{What the diagnosis LLM proposed at R0.} R0's per-question log $\mathcal{L}_0$ shows overall F1 of 0.336 with 75 zero-F1 questions, including 26 zero-F1 cases in Cat.~4 alone. The diagnosis module $\phi_{\text{diag}}$ reads $\mathcal{L}_0$ and emits the following priority actions (verbatim from the persisted trace):
\begin{quote}\small\itshape
1.~Enable semantic retrieval with \texttt{fusion\_mode='rrf'} and \texttt{semantic\_top\_k} in the low-mid range (12--16) so lexically-different but semantically-related memories (e.g., camping trip vs.\ Perseid meteor shower, painting descriptions) can be recalled.

2.~Increase retrieval depth and context breadth (\texttt{keyword\_top\_k} to $\sim$10--12, \texttt{max\_context} to $\sim$12--16) \textbf{especially for categories 4 and 5 via per\_category\_overrides} to fix the many abstentions and `not specified' failures for detailed episodic facts.

3.~Tighten and enrich extraction so specific concrete details are captured and retrievable.
\end{quote}
Notice that the diagnosis LLM names \emph{the exact failure mode} of this case (camping trip vs.\ Perseid meteor shower) inside priority action~1---a failure pattern it inferred from $\mathcal{L}_0$ alone, with no benchmark-specific cue.

\paragraph{Mechanism, round by round.}
\textit{(i)~R0$\to$R1 — recall.} BM25-only with $k{=}5$ retrieves the wrong camping memory (a separate ``Perseid meteor shower'' episode that shares the keyword \emph{camping}) and predicts a single tangential activity. The R1 update enables the semantic view (Eq.~\ref{eq:sem_cos}) and triples the context budget, so the answer LLM now sees \emph{all} camping-related episodes; the prediction expands to four activities, three of them correct, but the Perseid memory still leaks in—F1 jumps to 0.44.
\textit{(ii)~R1$\to$R2 — precision.} R2 raises the structured-view weight ($w_{\text{str}}{:}\,1.0\to1.8$) and turns on the recency factor $\text{rec}(m_i)$ (Eq.~\ref{eq:hybrid_score}); structured retrieval over extracted entities (Melanie, family, camping) re-anchors the top-ranked memories to the right episode, and the recency signal down-weights the older Perseid memory. The Perseid noise is dropped; F1 reaches 1.00.
\textit{(iii)~R2$\to$R3 — safety.} R2's aggressive expansion lifts this case but \emph{regresses overall F1 by $-$0.054}. The revert guard (Eq.~\ref{eq:meta_update}, $\tau_{\text{rev}}{=}0.01$) auto-rolls $\theta_{r+1}$ back to the best-so-far $\theta^{\star}{=}\theta_1$. The per-question prediction at R3 is essentially correct but missing the connector ``and'', costing 0.06 F1—a minor wording artifact rather than a content failure.
\textit{(iv)~R3$\to$R4 — polish.} The diagnosis LLM, reading R3's wording-only failures, proposes a per-category answer-style override for Cat.~4 that mandates explicit list connectors. Applied via the \texttt{per\_category\_overrides} mechanism (Eq.~\ref{eq:theta_space}), the connector is restored and F1 returns to 1.00.

\paragraph{Generalization.} This trace exemplifies a broader pattern. Across the 70 Cat.~4 probes in sample~0, the four configuration changes successively reduce zero-F1 cases ($26\to12\to11\to16\to9$, where the R3 uptick reflects the same revert artefact illustrated above) and lift the per-sample Cat.~4 F1 from 0.350 at R0 to 0.520 at R4 ($+$17\,pp). At the population level, the aggregate Cat.~4 trajectory across the full 10-sample evaluation climbs from 41.0\,\% at R0 to 49.6\,\% at R7 ($+$8.6\,pp), with each evolved dimension (multi-view fusion, structured and recency scoring, per-category answer styles, and answer verification) contributing a distinguishable share. The case study renders this aggregate effect at single-question resolution.

\section{Implementation Details}
\label{app:implementation}

\subsection{SQLite Schema}
\label{app:schema}

The memory store uses SQLite 3.35+ with FTS5 support. The core schema (version 6) includes:

\begin{itemize}[leftmargin=*,itemsep=1pt]
    \item \texttt{memories}: Primary storage table with columns for \texttt{memory\_id} (UUID), \texttt{scope\_id}, \texttt{memory\_type}, \texttt{content}, \texttt{summary}, \texttt{entities} (JSON), \texttt{topics} (JSON), \texttt{importance}, \texttt{confidence}, \texttt{reinforcement\_score}, \texttt{access\_count}, \texttt{embedding} (BLOB), \texttt{tags} (JSON), \texttt{status}, \texttt{supersedes} (JSON), \texttt{superseded\_by}, \texttt{expires\_at}, \texttt{created\_at}, \texttt{updated\_at}.
    \item \texttt{memories\_fts}: FTS5 virtual table indexing \texttt{content}, \texttt{summary}, \texttt{entities}, \texttt{topics} for efficient full-text search.
    \item \texttt{memory\_events}: Append-only audit log for all mutations.
    \item \texttt{memory\_links}: Relationship graph with typed edges (related, depends\_on, elaborates, contradicts).
    \item \texttt{schema\_version}: Migration tracking.
\end{itemize}

The database operates in WAL mode with normal sync and foreign keys enabled.

\subsection{Embedding Models}
\label{app:embeddings}

We support two embedding backends:

\begin{itemize}[leftmargin=*,itemsep=1pt]
    \item \textbf{HashingEmbedder}: A lightweight, deterministic hash-based embedder that maps tokens to dimensions via SHA-256 hashing. Produces $d{=}64$ dimensional vectors with $\ell_2$ normalization. Zero external dependencies; suitable for environments where installing ML libraries is impractical.
    \item \textbf{SentenceTransformerEmbedder}: Uses BAAI/bge-base-en-v1.5 (768-dim) from the \texttt{sentence-transformers} library. Provides semantic similarity for hybrid retrieval. Batch encoding with size 32 for efficiency.
\end{itemize}

All experiments use SentenceTransformerEmbedder.

\subsection{Efficiency Analysis}
\label{app:efficiency}

\paragraph{Self-evolution overhead.} A full 7-round evolution on one LoCoMo sample (200 QA, $\sim$900 memories) completes in 25--35\,min wall clock, dominated by QA evaluation LLM calls. Each round consists of index building ($\sim$5\,s per sample), QA evaluation ($\sim$15--20\,min for 200 questions with verification enabled), and LLM-powered diagnosis ($\sim$15\,s per sample). Convergence detection stops evolution automatically when the best-round metric plateaus.

\paragraph{Retrieval latency.} Multi-view index construction over $\sim$900 memories (SentenceTransformer encoding + BM25 index + metadata indices) completes in $\sim$5\,s. Per-query retrieval (semantic top-20 + BM25 top-8 + structured top-5 + entity-swap) averages 15\,ms, well within interactive requirements. Enabling answer verification adds one extra LLM call per question ($\sim$2--3\,s).

\paragraph{Storage and reproducibility.} The SQLite database with FTS5 index adds under 5\,MB per 1{,}000 memory units; extracted memory caches (JSON with structured metadata) average 150\,KB per LoCoMo sample. For every run we persist: (i)~per-round config snapshot, (ii)~complete \texttt{raw\_results.jsonl} containing every question, prediction, reference, every metric, and retrieved sources, (iii)~per-round summary, (iv)~the best-so-far configuration $\theta^\star$ snapshot, versioned alongside the code so that the autonomous trajectory is reproducible across runs, and (v)~extracted memory cache.

\section{Reproducibility}
\label{app:reproducibility}

\paragraph{Code.} Our implementation is available as a Python package with zero required external dependencies beyond the standard library and SQLite. Optional dependencies include \texttt{sentence-transformers} for semantic embeddings and an LLM API for memory extraction.

\paragraph{Compute.} All experiments were run on a single machine with an Apple M-series CPU (no GPU required for the memory system itself). LLM calls for extraction and diagnosis use GPT-5.1 via Azure OpenAI API. Answer generation uses GPT-4o for all categories. Self-evolution and consolidation are CPU-only operations.

\paragraph{Data.} LoCoMo \citep{maharana2024evaluating} is publicly available. MemBench \citep{membench2025} is publicly available.

\paragraph{Hyperparameters.} All evolvable hyperparameters and their valid ranges are listed in Table~\ref{tab:retrieval_config}. Default values were chosen based on preliminary experiments on a held-out validation set (2 LoCoMo samples) and remained fixed for all reported results.

\section{Prompt Catalog}
\label{app:prompts}

This appendix presents, verbatim, every LLM prompt used by \ours{}. Curly-brace placeholders (e.g., \texttt{\{context\}}, \texttt{\{question\}}) are substituted at runtime. Prompts are colour-coded by role: \textcolor{blue!70}{blue} = extraction / structural input, \textcolor{violet!70}{violet} = retrieval expansion, \textcolor{orange!80}{orange} = LoCoMo answer generation, \textcolor{teal!70!black}{teal} = MemBench answer generation, \textcolor{green!55!black}{green} = second-pass verification, \textcolor{yellow!50!brown}{yellow-brown} = diagnosis, and \textcolor{gray!70!black}{gray} = meta-evaluation.

\subsection{Extraction: Sliding-Window Memory Extraction}
\label{app:prompt_extract}

Called once per window $S^{(j)}$ of $W{=}40$ turns. The \{context\} slot receives the tail of the previous window's extractions to avoid duplication.

\begin{tcolorbox}[title={\textbf{Extraction: System + User Prompt}}, breakable, colback=blue!3, colframe=blue!40]
\small\ttfamily
You are a professional information extraction assistant.\\
Extract ALL valuable information from the following dialogue into structured memory entries.\\[0.3em]
\{context\}\\[0.3em]
{[Dialogue from \{date\}]}\\
\{dialogue\_text\}\\[0.3em]
{[Requirements]}\\
1. Complete Coverage: Generate entries for ALL facts, events, opinions, plans, feelings.\\
2. Force Disambiguation: PROHIBIT pronouns (he/she/it/they). Use actual names and absolute dates.\\
3. Lossless Restatement: Each entry must be complete, independent, self-contained.\\
4. Extract EVERY specific detail -- no paraphrasing fine-grained facts:\\
\ \ \ - Named entities: book/movie/song/game titles (keep quotation marks), brand names, places, pet names, nicknames, colors, specific activities, specific numbers.\\
\ \ \ - Quantities: exact counts, frequencies ("twice", "three times"), durations ("for 3 years", "since 2019").\\
\ \ \ - Lists: if someone mentions multiple items (books, hobbies, gifts), create ONE entry that lists them ALL.\\
\ \ \ - Gifts and possessions: for each gift/possession mentioned, create an explicit entry (who gave what to whom, when).\\
\ \ \ - Facts stated implicitly through dialogue: capture them as direct statements.\\
5. Cover names, places, objects, opinions, plans, feelings, events, dates, gifts, hobbies, relationships, pets, travel, food, books, art, music, work, family, health.\\[0.3em]
{[Output Format]}\\
Return a JSON array:\\
\lbrack\\
\ \ \{ "lossless\_restatement": "Complete sentence with all subjects, objects, time, location",\\
\ \ \ \ \ "keywords": ["keyword1", "keyword2"], "timestamp": "YYYY-MM-DD or null",\\
\ \ \ \ \ "location": "location or null", "persons": ["name1", "name2"],\\
\ \ \ \ \ "entities": ["entity1"], "topic": "topic phrase" \}\\
\rbrack\\[0.3em]
Return ONLY the JSON array. Extract at least 15 entries (more if the window contains multiple distinct facts). Prioritise completeness over brevity.
\end{tcolorbox}

\subsection{Retrieval Expansion: Query Decomposition}
\label{app:prompt_decompose}

Invoked when \texttt{enable\_query\_decomposition} is set (Eq.~\ref{eq:decomp}). \texttt{\{max\_n\}} is bound to \texttt{decomposition\_max\_subqs}.

\begin{tcolorbox}[title={\textbf{Query Decomposition: User Prompt}}, breakable, colback=violet!3, colframe=violet!40]
\small\ttfamily
Split this question into 1--\{max\_n\} single-hop sub-questions, each answerable from one piece of evidence. If the original question IS already single-hop, return only the original.\\[0.3em]
Question: \{question\}\\[0.3em]
Return JSON list of sub-question strings:\\
\lbrack "sub1", "sub2", ...\rbrack\\[0.3em]
Return ONLY the JSON array.
\end{tcolorbox}

\subsection{Answer Generation: LoCoMo}
\label{app:prompt_answer_locomo}

LoCoMo uses a category-aware adapter with three branches, all sharing one system message.

\begin{tcolorbox}[title={\textbf{LoCoMo: System Prompt (shared)}}, colback=orange!3, colframe=orange!40]
\small\ttfamily
Professional Q\&A assistant. Concise answers grounded in context. JSON output only.
\end{tcolorbox}

\begin{tcolorbox}[title={\textbf{LoCoMo Cat.~5 Adversarial: User Prompt}}, breakable, colback=orange!3, colframe=orange!40]
\small\ttfamily
Answer based on the context.\\[0.3em]
IMPORTANT: This question may deliberately swap person names. The CONTEXT contains the TRUE information; answer based on the context even if the question names seem off.\\[0.3em]
Question: \{question\}\\[0.3em]
Context:\\
\{context\}\\[0.3em]
Rules:\\
1. ALWAYS provide a substantive answer, never 'not specified'.\\
2. Answer in 1-5 words using exact facts from context.\\[0.3em]
Return JSON: \{"reasoning":"brief","answer":"concise"\}
\end{tcolorbox}

\begin{tcolorbox}[title={\textbf{LoCoMo Cat.~3 Inferential (base): User Prompt}}, breakable, colback=orange!3, colframe=orange!40]
\small\ttfamily
Question: \{question\}\\[0.3em]
Context:\\
\{context\}\\[0.3em]
This question asks for an INFERENCE or COUNTERFACTUAL judgement (e.g., 'Would X...', 'What would X likely...'). Your job is to synthesize a best-guess answer from the evidence, do NOT refuse.\\[0.3em]
Rules:\\
1. NEVER answer 'unknown' / 'not specified' / 'not mentioned'. The answer must always be a substantive judgement.\\
2. Answer in 1-6 words. Preferred forms:\\
\ \ \ - 'Would X...' -> 'Likely yes' / 'Likely no' / 'Yes' / 'No' (+ a short reason ONLY if very informative).\\
\ \ \ - 'What/Which would X...' -> name the most likely option.\\
3. Choose the option most consistent with the user's stated preferences, history, and values in the context.\\[0.3em]
Return JSON: \{"reasoning":"brief","answer":"concise"\}
\end{tcolorbox}

\begin{tcolorbox}[title={\textbf{LoCoMo Cat.~3 Nuanced-Inferential (discovered, 6 subtypes): User Prompt}}, breakable, colback=orange!3, colframe=orange!40]
\small\ttfamily
Gated by \texttt{locomo\_cat3\_inferential\_nuanced}. A regex classifier routes each Cat.~3 question to one of six subtypes; each receives a subtype-specific instruction block \texttt{\{specific\}} within a shared template:\\[0.3em]
Question: \{question\}\\
Context: \{context\}\\[0.3em]
\{specific\}\\[0.3em]
General rules:\\
1. CRITICAL: NEVER answer 'Unknown', 'Not specified', 'Not mentioned', empty string, or any refusal. Make the MOST PLAUSIBLE guess grounded in the speaker's stated preferences.\\
2. Prefer exact phrases from context over paraphrased abstract nouns ('Nintendo Switch', not 'a console').\\
3. For counterfactuals ('would X...'), pick the option most consistent with the speaker's stated preferences.\\
4. Geography: if the question asks for a STATE / COUNTRY and context only has a CITY / LANDMARK, infer the enclosing jurisdiction.\\
Return JSON: \{"reasoning":"brief","answer":"<answer>"\}\\[0.5em]
\normalfont\textbf{Subtype-specific instructions} (abbreviated):\\[0.3em]
\textbullet\ \textbf{Counting}: enumerate matching events, return spelled-out number (<5) or Arabic digit.\\
\textbullet\ \textbf{Location-hierarchy}: copy named jurisdiction if in context, else infer from city/landmark.\\
\textbullet\ \textbf{Open-ended synthesis}: 15--40 word paragraph with hedged claim + supporting details from context.\\
\textbullet\ \textbf{Either-or}: pick the one option most consistent with stated preferences (NOT 'Yes'/'No').\\
\textbullet\ \textbf{Yes/No}: default 'Yes'/'No'/'Likely yes'/'Likely no' in 1--3 words; optional clause only for strong counterfactual signal.\\
\textbullet\ \textbf{Multi-hop default}: copy distinctive noun phrase (1--6 words); optionally append a supporting clause (4--20 words total) when context contains a directly supporting fact.
\end{tcolorbox}

\begin{tcolorbox}[title={\textbf{LoCoMo Strict answer\_style (evolvable): User Prompt}}, breakable, colback=orange!3, colframe=orange!40]
\small\ttfamily
Question: \{question\}\\
Context: \{context\}\\[0.3em]
Rules:\\
1. Answer in 1-10 words. Use EXACT words/phrases from context.\\
2. Format conventions:\\
\ \ \ - 'how many/times' -> single Arabic numeral ('2', not 'two').\\
\ \ \ - 'when' / year questions -> 4-digit year ('2019') or 'YYYY-MM-DD' if date is known; never 'N years ago'.\\
\ \ \ - 'where' -> place name exactly as in context.\\
\ \ \ - 'what/who' -> shortest distinctive noun phrase in context.\\
3. NEVER answer 'Unknown', 'Not specified', 'Not mentioned'. Even if context is indirect, pick the single most plausible answer from what IS mentioned.\\
4. For multi-item questions (e.g. 'which movies'), list each item separated by a comma.\\[0.3em]
Return JSON: \{"reasoning":"brief","answer":"concise"\}
\end{tcolorbox}

\subsection{Answer Generation: MemBench (MCQ)}
\label{app:prompt_answer_membench}

MemBench is multiple-choice.

\begin{tcolorbox}[title={\textbf{MemBench: System Prompt}}, colback=teal!3, colframe=teal!50]
\small\ttfamily
You are a memory-grounded multiple-choice question answerer. You MUST pick exactly ONE letter (A/B/C/D). JSON only.
\end{tcolorbox}

\begin{tcolorbox}[title={\textbf{MemBench: User Prompt Template}}, breakable, colback=teal!3, colframe=teal!50]
\small\ttfamily
Question: \{question\}\\[0.3em]
Options:\\
\ \ A) \{opt\_a\}\\
\ \ B) \{opt\_b\}\\
\ \ C) \{opt\_c\}\\
\ \ D) \{opt\_d\}\\[0.3em]
Context (memory snippets):\\
\{context\}\\[0.3em]
Rules:\\
1. Pick EXACTLY one letter from \{A,B,C,D\}.\\
2. Base your answer on the context; if context is incomplete still pick the most plausible option.\\
3. Return JSON: \{"reasoning":"brief","answer":"X"\} where X is a single letter.
\end{tcolorbox}

\subsection{Answer Verification (Second Pass)}
\label{app:prompt_verify}

Invoked when \texttt{enable\_answer\_verification} is set (Eq.~\ref{eq:verify}).

\begin{tcolorbox}[title={\textbf{Verifier: System Prompt}}, colback=green!3, colframe=green!40]
\small\ttfamily
Answer verifier. JSON output only.
\end{tcolorbox}

\begin{tcolorbox}[title={\textbf{Verifier Strict (default): User Prompt}}, breakable, colback=green!3, colframe=green!40]
\small\ttfamily
Question: \{question\}\\[0.3em]
Context:\\
\{context\}\\[0.3em]
Candidate answer: \{candidate\}\\[0.3em]
Review the candidate answer. If it says 'Unknown' or 'Not specified', replace it with the most likely answer from the context. Format numbers as Arabic digits, years as YYYY. Keep the answer concise (1-8 words).\\[0.3em]
Return JSON: \{"reasoning":"brief","answer":"final answer"\}
\end{tcolorbox}

\begin{tcolorbox}[title={\textbf{Verifier Multi-Candidate: Instruction Swap}}, colback=green!3, colframe=green!40]
\small\ttfamily
Review the candidate answer. If it is wrong, give the correct answer. If the candidate is 'Unknown' but the context contains any relevant fact, pick the most plausible option. Consider 2-3 candidate answers and pick the best.
\end{tcolorbox}

\subsection{Diagnosis: LLM-Powered Failure Analysis}
\label{app:prompt_diagnosis}

Invoked once per evolution round over the per-question raw log $\mathcal{L}_r$ (Eq.~\ref{eq:raw_log}).
Returns the structured proposal.

\begin{tcolorbox}[title={\textbf{Diagnosis Engine: System + User Prompt}}, breakable, colback=yellow!5, colframe=yellow!50!brown]
\small\ttfamily
You are the diagnosis engine of a self-evolving memory system. Your job is to turn evaluation failures into a concrete next-round action that moves the system toward SOTA.\\[0.3em]
\#\# System Info\\
- Benchmark: \{benchmark\}\\
- Total memories: \{total\_memories\}\\
- Total questions: \{total\_questions\}\\
- Overall score: \{overall\_f1:.4f\}\\
- Zero-score count: \{zero\_count\}/\{total\_questions\}\\
- Current config (JSON): \{current\_config\}\\[0.3em]
\#\# Failure Analysis Summary\\
\{failure\_summary\}\\[0.3em]
\#\# Per-Category Breakdown\\
\{category\_breakdown\}\\[0.3em]
\#\# Sample Failures (worst cases)\\
\{sample\_failures\}\\[0.3em]
\#\# Tier-1 TODO checklist: levers that are still OFF in the incumbent\\
\{todo\_checklist\}\\[0.3em]
\textit{(Dynamically generated: lists each disabled lever whose symptom is present in the current failure data. Prefer picking ONE item per round until empty.)}\\[0.3em]
chunk\_size\_on\_failure, min\_restatement\_words.\\[0.3em]
\#\# Decision Rubric\\
1. If many 'abstention' failures -> raise top\_k, widen max\_context, consider rrf fusion.\\
2. If many 'wrong answer' failures with high retrieval -> lower max\_context or raise weights for strongest view.\\
3. If temporal category weakness -> enable time\_decay\_half\_life\_days.\\
4. If adversarial category weakness -> enable\_entity\_swap=true.\\
5. If multi-hop weakness -> reflection\_rounds >= 1.\\
6. If ONE category lags -> per\_category\_overrides (preserve gains elsewhere).\\
7. Prefer enabling something disabled BEFORE tuning a small int.\\
8. If residual 'Unknown' or format-mismatch -> enable\_answer\_verification=true.\\
9. LoCoMo prompt-surface flags are highest-ROI when their symptom matches; propose them early.\\[0.3em]
\#\# Output\\
Return JSON with `parameter\_suggestions` as a flat dict of field -> new value. Fields MUST match RetrievalConfig field names exactly. Only include fields you want to change.\\[0.3em]
\{\{\\
\ \ "root\_causes": \{\{"extraction\_gap": \{\{...\}\}, "retrieval\_miss": \{\{...\}\}, "answer\_error": \{\{...\}\}\}\},\\
\ \ "missing\_topics": ["topic1", "topic2"],\\
\ \ "parameter\_suggestions": \{\{"fusion\_mode": "rrf", "semantic\_top\_k": 15\}\},\\
\ \ "extraction\_suggestions": \{\{"window\_size": 30\}\},\\
\ \ "per\_category\_proposals": \{\{"5": \{\{"enable\_entity\_swap": true\}\}\}\},\\
\ \ "priority\_actions": ["action1", "action2", "action3"]\\
\}\}\\[0.3em]
Return ONLY JSON.
\end{tcolorbox}